\documentclass[11pt]{article}

\usepackage[preprint]{acl}

\usepackage{times}
\usepackage{latexsym}

\usepackage{amsmath}
\usepackage{amssymb}

\usepackage[T1]{fontenc}

\usepackage[utf8]{inputenc}

\usepackage{microtype}

\usepackage{inconsolata}
\usepackage{booktabs}
\usepackage{array}

\usepackage{graphicx}

\title{Harnessing Language Models as Optimization Modeling Agents}

\author{
  \textbf{Yunxiang Luo\textsuperscript{1}\thanks{Equal contribution.}},
  \textbf{Xin Xu\textsuperscript{1}\footnotemark[1]},
  \textbf{Tianhao Chen\textsuperscript{1}\footnotemark[1]},
  \textbf{Zhixi Xiong\textsuperscript{1}}, 
  \textbf{Peiqi Jiang\textsuperscript{1}},
  \\
  \textbf{Fan Zhang\textsuperscript{1}},
  \textbf{Wei Shao\textsuperscript{2}},
  \textbf{Xuejian Gao\textsuperscript{2}},
  \textbf{Yuwei Fan\textsuperscript{2}},
  \textbf{Can Yang\textsuperscript{1,}\thanks{
    Correspondence:
    \href{mailto:macyang@ust.hk}{macyang@ust.hk}
  }}
  \\
  \\
  \textsuperscript{1}The Hong Kong University of Science and Technology,
  \\
  \textsuperscript{2}Huawei
}

\begin{document}
\maketitle
\begin{abstract}
Large language models (LLMs) can translate natural-language optimization problems into solver programs, but reliably producing correct formulations and implementations remains challenging. Existing approaches improve optimization modeling through agentic workflows or task-specific post-training. This raises a practical question: can a lightweight harness using a single model role achieve competitive performance without additional training? We introduce \textsc{OptiCycle} and \textsc{OptiState}, two lightweight
inference harnesses for optimization modeling, and evaluate them on nine benchmarks comprising 2,634 problems using
11 Qwen variants spanning 0.6B--27B parameters.  
Our results show that execution-guided revision consistently
improves macro accuracy over direct generation for all evaluated
models, with gains of up to 49.46 percentage. Despite the simplicity, both harnesses outperform
six adapted agentic workflows under same model comparisons
and four evaluated post-trained optimization models at comparable
model sizes without additional training. Explicit formulation
state provides further but model-dependent improvements,
suggesting that additional structure is not universally
beneficial. 
\end{abstract}

\begin{figure*}[t]
\centering
\includegraphics[width=\textwidth]{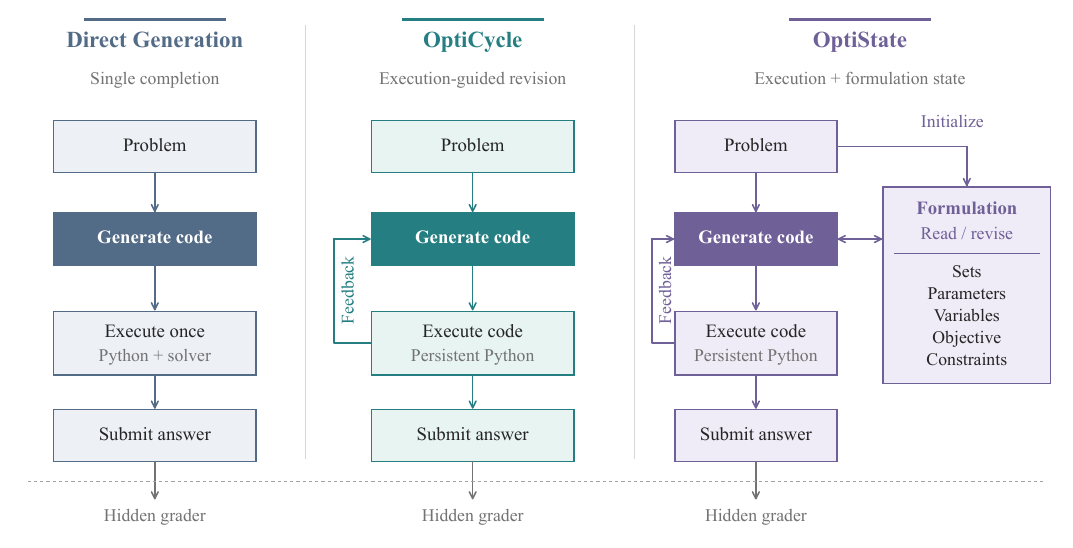}
\caption{Three inference interfaces for the same frozen model.
OptiCycle adds execution feedback and persistent Python state;
OptiState additionally maintains a revisable five-element formulation.
The hidden grader evaluates submitted answers and provides no feedback
during inference.}
\label{fig:method_overview}
\end{figure*}

\section{Introduction}

Operations research (OR) provides a \mbox{principled} foundation for decision-making in resource-constrained settings, with optimization models \mbox{supporting} applications such as supply-chain \mbox{management} and transportation \cite{dantzig2016linear,hillier1995introduction}. In practice, an optimization problem often begins as a natural-language \mbox{description} of the decision context, available \mbox{resources}, and desired outcomes. Translating this description into a correct optimization model requires identifying decision variables, an objective, and constraints while preserving the semantics of the original problem \cite{ramamonjison2023nl4opt}. 
Natural-language optimization modeling therefore connects language understanding, mathematical formulation, and executable problem solving.

Large language models (LLMs) offer a way to automate this translation from problem descriptions to executable optimization models. Existing approaches pursue two complementary directions. One organizes inference through specialized agents or stages for formulation, coding, and debugging \cite{xiao2024chain,ahmaditeshnizi2024optimus}. The other treats it as a reasoning task and develops domain-specific capabilities through \mbox{supervised} fine-tuning (SFT) or reinforcement learning (RL) \cite{huang2025orlm,chen2026solver}. The two \mbox{directions} differ in whether they modify the \mbox{inference} workflow or the model parameters.

Agentic post-training further connects model \mbox{capabilities} with the systems surrounding them: models can be trained to use tools and interact with environments \cite{kimi2025k2,glm2025arc}, but the opportunities to exercise these \mbox{capabilities} depend on the inference procedure. The same frozen LLM can act as a direct generator or repeatedly execute and revise a solver program. An \emph{inference harness} is the layer that organizes an LLM's tools, context, state, and control flow \cite{meng2026harness}. Comparisons across optimization systems are difficult to interpret when backbones, benchmarks, inference procedures, and grading protocols all differ. This motivates our central question: \emph{how do different harnesses shape the optimization-modeling performance of a frozen LLM, and when does explicit formulation state add value?}

In this paper, we study this question through a unified harness view of optimization modeling (\mbox{Figure}~\ref{fig:method_overview}). Direct Generation is the single-step \mbox{setting}: the model produces a solver program \mbox{without} an opportunity to revise it after execution. We design two OR-oriented harnesses for interactive modeling. \textsc{OptiCycle} organizes code generation, solver execution, feedback interpretation, and \mbox{revision} into a cycle within a persistent Python environment. \textsc{OptiState} adds a structured and revisable formulation comprising sets, parameters, variables, the objective, and constraints. This state makes intermediate modeling decisions explicit and available for reconsideration during subsequent interaction. Both harnesses use a single model role with frozen weights; reference answers remain \mbox{hidden}, and grading occurs only after submission.

We evaluate these settings on nine benchmarks comprising 2,634 problems and 11 Qwen \mbox{variants} from 0.6B to 27B parameters. \textsc{OptiCycle} \mbox{improves} \texttt{Qwen3.5-9B} macro accuracy by 18.74 percentage points over Direct Generation. In a shared-backbone comparison with six adapted OR workflows, it achieves 62.74\% on \texttt{Qwen3-4B} and 65.21\% on \texttt{Qwen3.5-9B}, exceeding the strongest competing baselines by 8.53 and 4.29 percentage points, respectively. These comparisons share the backbone and evaluation protocol while retaining method-specific inference budgets. 

Our contributions are as follows:

\begin{itemize}
    \item A unified harness view of optimization modeling for LLMs. We compare direct generation, execution-guided revision, and formulation-state-guided revision while keeping model weights fixed.

    \item Two simple and effective modeling harnesses. \textsc{OptiCycle} enables iterative code execution and repair, while \textsc{OptiState} adds a revisable representation of the optimization \mbox{formulation}.

    \item Comprehensive evaluation across diverse benchmarks and baselines. We evaluate 11 Qwen variants on nine benchmarks \mbox{containing} 2,634 problems and compare against six adapted workflows under shared backbones and grading.

    \item An analysis of effectiveness and cost. Execution-guided revision produces broad gains, whereas formulation state provides model-dependent benefits at higher token \mbox{consumption}.
\end{itemize}

\section{Related Work}
\subsection{LLMs for Optimization Modeling}
Recent work on LLM-based optimization \mbox{modeling} broadly follows two methodological directions: \mbox{improving} inference-time modeling procedures and adapting model parameters specifically for optimization tasks.

\paragraph{Agent Systems for OR.}
One direction \mbox{improves} optimization modeling through task-specific prompting and structured inference workflows. Existing systems organize formulation, coding, execution and debugging into coordinated stages or specialized roles \cite{deng2024cafa,xiao2024chain,ahmaditeshnizi2024optimus,monteiro2026optiagent}. Beyond stage decomposition, recent approaches explore formulation search, execution-aware verification, iterative repair, and the reuse of correction experiences \cite{astorga2024autoformulation,wang2025ormind,song2026nemo,xie2026experience}. These methods improve the inference process without necessarily changing the underlying model. Our work focuses on this setting, studying OR-oriented harnesses that support iterative modeling within a single agent role while holding model weights fixed.

\paragraph{Reasoning Models for Optimization Modeling.}
A complementary direction develops optimization-specific reasoning through parameter adaptation. Data synthesis and SFT provide supervision for \mbox{formulation} and code generation \cite{huang2025orlm,wang2024optibench,lu2025optmath,wu2025training}, while solver-grounded RL and \mbox{process} \mbox{supervision} further develop optimization \mbox{reasoning} \cite{chen2026solver,xiao2026deepor,tang2026calm,guan2026execution}. Related approaches \mbox{extend} parameter adaptation to test time \cite{ding2026or,li2026staror}. Our study keeps model weights fixed. In particular, \textsc{OptiState} draws on LLMOPT's five-element formulation \cite{shu2025llmopt} as a revisable inference-time state rather than training supervision.

\paragraph{Benchmarks for Optimization Modeling.}
Optimization-modeling benchmarks have \mbox{progressively} expanded in scope and difficulty. NL4Opt focuses on natural-language linear programming problems \cite{ramamonjison2023nl4opt}, while NLP4LP covers both linear and mixed-integer programming \cite{ahmaditeshnizi2024optimus}. MAMO and ComplexOR broaden the range of modeling tasks and problem \mbox{structures} \cite{huang2024mamo,xiao2024chain}. IndustryOR, OptiBench, and OptMATH further extend the evaluation to more diverse domains, formulations, and levels of complexity \cite{huang2025orlm,wang2024optibench,lu2025optmath}. We use existing optimization-modeling tasks as a controlled testbed for comparing inference-time harness designs.

\subsection{Agentic Language Models and Harnesses}
LLM-based agents have developed through both \mbox{externally} organized workflows and improvements in model capabilities. Approaches such as \mbox{AutoGen} and MetaGPT coordinate \mbox{specialized} roles, while ReAct and Reflexion connect \mbox{reasoning} with actions and feedback without \mbox{updating} model weights \cite{wu2023autogen,hong2024metagpt,yao2023react,shinn2023reflexion}. More recently, agentic RL incorporates tool use and environment interaction into model post-training \cite{kimi2025k2,glm2025arc}. \mbox{Evaluating} the resulting capabilities calls for both task-level benchmarks and an explicit account of the harness used at inference time.

\paragraph{Agentic Benchmarks.}
General-purpose evaluations assess software maintenance, terminal tasks, and program reconstruction, as exemplified by SWE-bench, Terminal-Bench, and ProgramBench \cite{jimenez2024swe,merrill2026terminal,yang2026programbench}. Within OR, recent benchmarks \mbox{evaluate} end-to-end workflows and solver-guided model repair \cite{li2026oragentbench,ao2026orloopbench}. Related combinatorial optimization benchmarks focus on algorithm and heuristic design \cite{sun2025cobench,chen2026heurigym}. We use existing optimization modeling tasks to investigate how harnesses affect frozen LLMs under a shared evaluation protocol.

\paragraph{Inference Harnesses.}
A harness organizes the tools, context, state, and control flow surrounding an LLM \cite{meng2026harness}. Coding-agent systems such as Claude Code, Codex CLI, and \mbox{OpenCode} provide implementations of this inference-time layer, whereas Harbor provides shared \mbox{evaluation} interfaces for running agents across tasks and \mbox{environments} \cite{anthropic2026code,openai2026codex,opencode2026docs,harbor2026motivation}. Recent research examines how harness choices affect measured \mbox{performance} and how harnesses can be composed and adapted \cite{yao2026harnessbench,chen2026harnessx}. We investigate this design space for optimization \mbox{modeling} through \textsc{OptiCycle} and \textsc{OptiState}, treating Direct Generation as the single-step \mbox{special} case.

\section{A Unified Harness View}
\label{sec:method}

We organize optimization modeling around a frozen language model and a
harness that \mbox{controls} its execution environment, interaction \mbox{history},
and persistent state. Within this view, Direct \mbox{Generation} is the
single-step setting, \mbox{\textsc{OptiCycle}} \mbox{introduces} execution-guided
revision, and \mbox{\textsc{OptiState}} additionally exposes a revisable
formulation (Figure~\ref{fig:method_overview}). Both of our harnesses use
one model role and keep its weights fixed \mbox{throughout} inference.

\begin{table*}[!t]
\centering

\footnotesize
\setlength{\tabcolsep}{1.5pt}
\renewcommand{\arraystretch}{1.12}
\caption{Benchmark accuracy (\%) for Direct Generation, adapted agent baselines, and our inference harnesses with shared backbones. Values are avg@$k$; Macro is the unweighted mean over nine benchmarks. Bold and underline mark the highest and second-highest values within each backbone and column, respectively, including ties.}
\label{tab:agent_macro}
\begin{tabular*}{\textwidth}{@{}m{68pt}*{10}{|>{\centering\arraybackslash}m{\dimexpr(\textwidth-68pt-20\tabcolsep-10\arrayrulewidth)/10\relax}}@{}}
\toprule
Method & \textbf{\shortstack{Complex\\OR}} & \textbf{\shortstack{Industry\\OR}} & \textbf{\shortstack{Opti\\Bench}} & \textbf{\shortstack{MAMO\\easy}} & \textbf{\shortstack{MAMO\\complex}} & \textbf{NL4Opt} & \textbf{NLP4LP} & \textbf{OptMath} & \textbf{\shortstack{ICML\\Comp.}} & \textbf{Macro} \\
\cmidrule(lr){2-11}
& avg@64 & avg@16 & avg@4 & avg@4 & avg@8 & avg@8 & avg@8 & avg@8 & avg@8 & Avg. \\
\midrule
\multicolumn{11}{c}{\textbf{Qwen3-4B}} \\
\midrule
Direct Generation & 29.08 & 33.88 & 54.38 & 80.06 & 27.61 & 72.45 & 72.88 & 9.41 & 72.62 & 50.26 \\
\addlinespace[3pt]
CAFA & 46.09 & 35.06 & 60.79 & 79.49 & 30.15 & 79.24 & 69.16 & 11.14 & 76.77 & 54.21 \\
Reflexion & 43.84 & 25.50 & 50.99 & 82.52 & 13.27 & 77.17 & 70.71 & 7.15 & 70.52 & 49.07 \\
Chain-of-Experts & 28.30 & 23.31 & 48.84 & 71.09 & 6.93 & 71.25 & 72.68 & 4.14 & 65.12 & 43.52 \\
OptiMUS & 47.92 & 29.94 & 57.77 & \underline{85.70} & 15.23 & 76.09 & \textbf{76.55} & 10.54 & 73.20 & 52.55 \\
AutoFormulation & 42.80 & 32.25 & 57.64 & 73.66 & 20.62 & 73.37 & 69.63 & 11.22 & 69.51 & 50.08 \\
ORMind & 51.74 & 33.88 & 55.62 & 83.90 & 15.58 & 72.66 & \underline{75.77} & 8.89 & 70.52 & 52.06 \\
\addlinespace[3pt]
OptiCycle & \textbf{66.58} & \underline{40.06} & \underline{70.12} & \underline{85.70} & \underline{33.06} & \underline{86.25} & 74.64 & \textbf{23.04} & \underline{85.18} & \underline{62.74} \\
OptiState & \underline{65.45} & \textbf{40.13} & \textbf{73.18} & \textbf{87.58} & \textbf{36.49} & \textbf{89.46} & 75.67 & \underline{21.76} & \textbf{88.23} & \textbf{64.22} \\
\midrule
\multicolumn{11}{c}{\textbf{Qwen3.5-9B}} \\
\midrule
Direct Generation & 34.64 & 27.50 & 51.20 & 69.13 & 22.22 & 71.79 & 62.86 & 11.37 & 67.53 & 46.47 \\
\addlinespace[3pt]
CAFA & 45.05 & 34.06 & 59.09 & 73.62 & 24.17 & 71.85 & 65.55 & 12.20 & 72.93 & 50.95 \\
Reflexion & 54.69 & 39.00 & 69.75 & 84.93 & \underline{32.94} & \underline{88.21} & 71.64 & 22.44 & 84.70 & 60.92 \\
Chain-of-Experts & 44.62 & 36.06 & 58.22 & 70.94 & 22.51 & 72.72 & 68.70 & 15.74 & 73.41 & 51.44 \\
OptiMUS & 55.30 & 40.50 & 68.80 & 85.74 & 29.56 & 82.17 & 77.32 & 24.70 & 81.95 & 60.67 \\
AutoFormulation & 45.57 & 34.00 & 61.12 & 69.17 & 24.64 & 73.80 & 69.83 & 20.48 & 74.45 & 52.56 \\
ORMind & 46.70 & 37.94 & 56.86 & 76.96 & 27.73 & 62.66 & 58.68 & 16.94 & 65.06 & 49.95 \\
\addlinespace[3pt]
OptiCycle & \underline{64.06} & \underline{43.62} & \underline{73.68} & \textbf{89.00} & 31.93 & 87.66 & \textbf{77.79} & \underline{31.40} & \underline{87.74} & \underline{65.21} \\
OptiState & \textbf{65.54} & \textbf{47.19} & \textbf{75.33} & \underline{86.66} & \textbf{38.39} & \textbf{92.66} & \underline{77.58} & \textbf{33.81} & \textbf{90.49} & \textbf{67.52} \\
\bottomrule
\end{tabular*}

\end{table*}

\subsection{Task and Harness Interface}

Given a natural-language problem $q$ and its \mbox{numerical} data, the model
must infer decision \mbox{variables} and their domains, the objective, and the
constraints, then implement and solve the \mbox{formulation} as a solver
program. The \mbox{submitted} answer is an optimal objective value or a \mbox{status}
in $\{\texttt{infeasible},\texttt{unbounded}\}$. Missing or \mbox{unparseable}
submissions count as failures.

The harness separates the model's proposed \mbox{computation} from its
execution. The model \mbox{produces} code, the environment runs it, and the
controller \mbox{determines} what feedback and state are available for the next
call and when to stop. Direct Generation makes one model call to produce
a complete solver program and executes it once. It has no opportunity to
revise the program after observing its behavior. Our interactive
harnesses maintain execution state across calls and provide summaries
of recent actions and observations, allowing the same model to revise
its work.

\subsection{Iterative Refinement through Execution Feedback}
\label{sec:local_agent}

\textsc{OptiCycle} organizes optimization modeling as a sequence of
code generation, execution, \mbox{observation}, and revision. The model can
build a solver program incrementally, inspect its behavior, and choose
its next operation from the execution \mbox{results}. Each trajectory runs in
a persistent Python interpreter, so imported modules, variables, and
solver objects remain available across turns. This continuity lets the
model resume from its current work: it can repair an indexing error,
modify a \mbox{constraint}, or rerun a solver using objects it has already
created.

The harness separates persistent execution state from the context
presented to the model. The \mbox{interpreter} retains live objects, workspace
files store intermediate artifacts, and the prompt \mbox{combines} the
original problem with a bounded summary of recent actions and execution
feedback. These summaries record the purpose of each action and the
resulting output or exception. Together with optimization-specific
modeling instructions and recovery hints, they guide the next code
operation while keeping the interaction context compact.

Each action $a_t$ specifies a purpose and code $c_t$. The loop proceeds as
\begingroup
\setlength{\abovedisplayskip}{9pt}
\setlength{\belowdisplayskip}{9pt}
\setlength{\abovedisplayshortskip}{6pt}
\setlength{\belowdisplayshortskip}{9pt}
\begin{equation}
\label{eq:local_execution}
\begin{aligned}
    &a_t \sim M_{\theta}(\cdot \mid q,\mathcal{C}(H_t)), \\[3pt]
    &(s_{t+1},o_t) = \mathcal{E}(c_t,s_t), \\[3pt]
    &H_{t+1} = H_t \mathbin{\Vert} (a_t,o_t).
\end{aligned}
\end{equation}
\endgroup
\noindent
where $s_t$ is the persistent execution state, $o_t$ is the execution
feedback, and $\mathcal{C}$ constructs the bounded context from the
recorded history $H_t$. The operator $\Vert$ appends the latest action
and observation. The loop ends upon valid submission or exhaustion
of the execution budget.

\subsection{Explicit Formulation as Revisable State}
\label{sec:five_element}

\textsc{OptiState} extends this design by making the model's
interpretation of the optimization \mbox{problem} an explicit part of the
workspace. A \mbox{separate} \mbox{formulation} records the decisions that guide
\mbox{implementation:} which entities are indexed, which \mbox{quantities} are known,
what can be chosen and how the objective and constraints relate to these
\mbox{quantities}. Keeping these decisions together gives the model a
structured reference as it develops the solver \mbox{program}.

Before code execution begins, an initial model call constructs a
five-element formulation, \mbox{following} LLMOPT~\cite{shu2025llmopt}:
\begin{equation}
\label{eq:five_element}
    z_0 = (z^{\mathrm{set}},z^{\mathrm{param}},z^{\mathrm{var}},
           z^{\mathrm{obj}},z^{\mathrm{con}}).
\end{equation}
The components describe sets and indices, known parameters, decision
variables and their domains, the objective, and constraints. The
formulation is saved in the workspace and supplied as a \mbox{context} for
code generation. The model translates this \mbox{representation} into solver
operations within the same persistent execution loop as
\textsc{OptiCycle}.

The formulation remains editable throughout the trajectory. The model
can inspect the stored representation, update its variable definitions
or constraints, and revise the corresponding solver code. For example,
a change in how a quantity is indexed can be recorded in the formulation
and carried through to variable creation and constraint construction.
This makes formulation part of the ongoing modeling process, extending
the agent's working state beyond Python objects and execution history.
\textsc{OptiState} retains the execution feedback and submission workflow
of \textsc{OptiCycle}, with an additional check for a parseable
formulation \mbox{containing} the required fields at submission.

\section{Experiments}
\label{sec:experiments}

\subsection{Experimental Settings}
\label{sec:experimental_setup}

We summarize the baselines and evaluation \mbox{protocols} here. Samples Per-benchmark
, \mbox{workflow} budgets, and execution details are provided in
Appendix~\ref{sec:experimental_details}.

\paragraph{Baselines.}
We compare \textsc{OptiCycle} and \mbox{\textsc{OptiState}} with Direct
Generation across 11 frozen Qwen variants spanning 0.6B--27B \mbox{parameters}.
On \texttt{Qwen3-4B} and \texttt{Qwen3.5-9B}, we additionally compare
with CAFA~\cite{deng2024cafa},
Reflexion~\cite{shinn2023reflexion},
\mbox{Chain-of-Experts}~\cite{xiao2024chain},
\mbox{OptiMUS}~\cite{ahmaditeshnizi2024optimus},
AutoFormulation~\cite{astorga2024autoformulation}, and
ORMind~\cite{wang2025ormind}.
Adaptations and method-specific budgets are detailed in
Appendix~\ref{sec:baseline_details}.

\paragraph{Evaluation settings.}
We evaluate on nine \mbox{benchmarks }comprising 2,634 problems and report
avg@$k$ (average sampled-trajectory accuracy) and its unweighted macro
average across benchmarks. Missing scores count as failures.
All model calls \mbox{allow} up to 16,384 output
tokens. Dataset, \mbox{sampling}, execution, and grading details are \mbox{provided}
in \mbox{Appendix}~\ref{sec:experimental_details}.

\subsection{Gains from Executable Interaction}
\label{sec:results_scaling}
\begin{figure*}[!t]
\centering
\includegraphics[width=\textwidth]{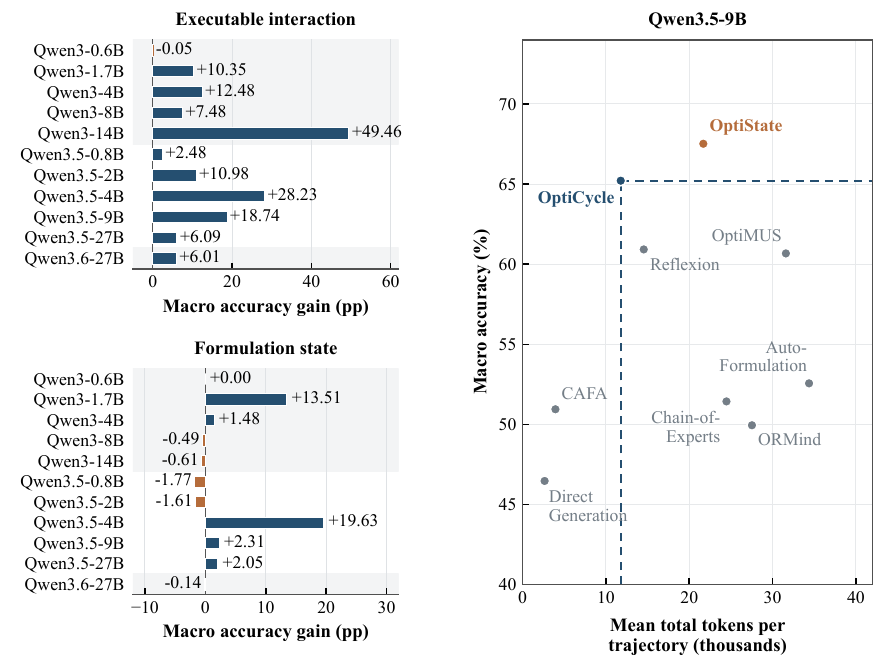}
\caption{Accuracy gains and token consumption across nine benchmarks.
Left, upper: OptiCycle gains over direct generation; left, lower:
OptiState gains over OptiCycle, in percentage points. Background bands
separate model series; the gain panels use different horizontal scales
and differences between rounded scores. Right: macro accuracy versus mean
recorded input plus output tokens per trajectory on Qwen3.5-9B, with
both metrics averaged equally across benchmarks. Dashed lines mark
OptiCycle's accuracy and token consumption.}
\label{fig:harness_gains}
\label{fig:accuracy_tokens}
\end{figure*}
\begin{figure*}[t]
\centering
\includegraphics[width=\textwidth]{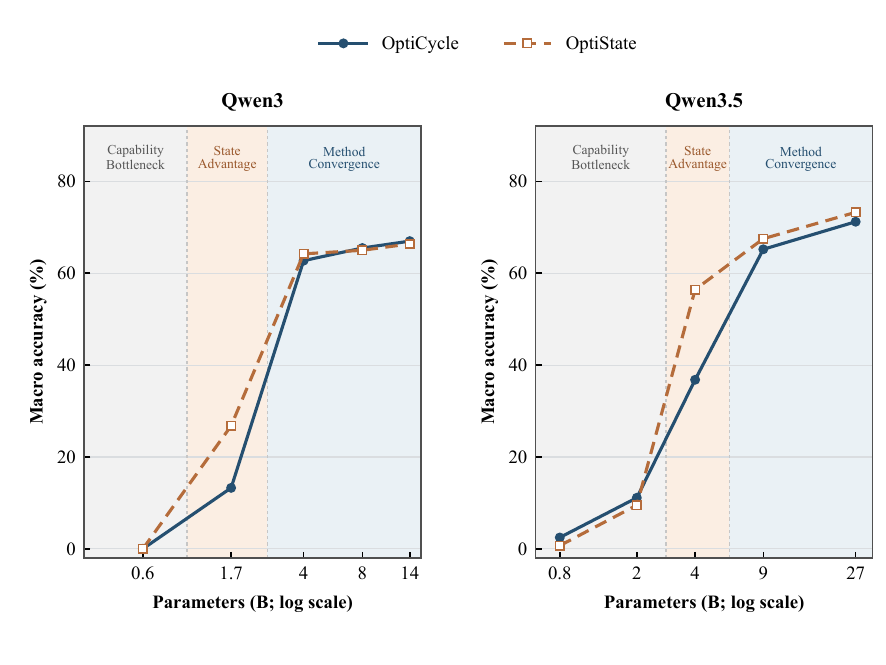}
\caption{Model-size scaling within the Qwen3 (left) and Qwen3.5 (right)
families. Model size is shown in
billions of parameters on a logarithmic scale; accuracy is the macro
average across nine benchmarks. Blue solid lines with circles denote
OptiCycle; orange dashed lines with squares denote OptiState. Shaded
bands mark three qualitative regimes: Capability Bottleneck, State
Advantage, and Method Convergence.}
\label{fig:model_scaling}
\end{figure*}

Table~\ref{tab:scale_results} shows that \textsc{OptiCycle} consistently
\mbox{improves} macro accuracy over direct generation for the evaluated
models above one billion \mbox{parameters}, whereas the smallest models
remain weak under all three methods. Figure~\ref{fig:model_scaling}
shows that harness {accuracy} generally increases with model size within
each family, with a sharper rise at intermediate scales and smaller
gains among the larger variants. Executable interaction thus benefits
a broad range of backbones, but does not eliminate the performance
gap between model scales.
\begin{table}[t]
\centering

\small
\caption{
Macro-average accuracy (\%) across nine optimization benchmarks.
 $\dagger$ marks evaluations with missing scores counted as failures; see Appendix~\ref{sec:sampling_details}.}
\label{tab:scale_results}
\begin{tabular}{lccc}
\toprule
Model & direct & OptiCycle & OptiState \\
\midrule
Qwen3-0.6B & \textbf{0.06\%} & 0.01\% & 0.01\% \\
Qwen3-1.7B & 2.93\% & 13.28\% & \textbf{26.79\%} \\
Qwen3-4B & 50.26\% & 62.74\% & \textbf{64.22\%} \\
Qwen3-8B & 58.01\% & \textbf{65.49\%} & 65.00\% \\
Qwen3-14B    & 17.53\% & \textbf{66.99\%} & 66.38\% \\
\midrule
Qwen3.5-0.8B & 0.00\% & \textbf{2.48}\% & 0.71\% \\
Qwen3.5-2B & 0.15\% & \textbf{11.13}\% & 9.52\%$^{\dagger}$ \\
Qwen3.5-4B & 8.58\% & 36.81\% & \textbf{56.44}\% \\
Qwen3.5-9B & 46.47\% & 65.21\% & \textbf{67.52}\% \\
Qwen3.5-27B & 65.12\% & 71.21\%$^{\dagger}$ & \textbf{73.26}\% \\
Qwen3.6-27B  & 66.47\% & \textbf{72.48\%} & 72.34\% \\
\bottomrule
\end{tabular}

\end{table}

\subsection{Comparison with Adapted Agent Baselines}
\label{sec:results_agents}
Both \textsc{OptiCycle} and \textsc{OptiState} achieve higher macro
accuracy than all six adapted \mbox{baselines} on Qwen3-4B and Qwen3.5-9B
(Table~\ref{tab:agent_macro}). \mbox{\textsc{OptiState}} achieves the highest
macro accuracy in both comparisons.

\raggedbottom
\subsection{Accuracy and Token Consumption}
\label{sec:results_tokens}

On \texttt{Qwen3.5-9B}, \textsc{OptiCycle} achieves higher macro
accuracy with fewer recorded tokens than Reflexion, Chain-of-Experts,
OptiMUS, \mbox{AutoFormulation}, and ORMind (Figure~\ref{fig:accuracy_tokens}, right).
It uses 11.77 thousand input plus output tokens per trajectory,
versus 14.54 for Reflexion, 31.60 for \mbox{OptiMUS}, and 34.38 for
AutoFormulation. \mbox{Compared} with Reflexion, it gains 4.29 \mbox{percentage}
points in accuracy while using 19.0\% fewer tokens. Direct Generation
and CAFA use fewer tokens but achieve lower accuracy.
\textsc{OptiState} adds 2.31 percentage points over \textsc{OptiCycle}
at 84.2\% higher token consumption. See
Appendix~\ref{sec:experimental_details} for token accounting.

\subsection{Comparison with Post-Trained Models}
\label{sec:released_models}
\begin{figure}[t]
\centering
\includegraphics[width=\columnwidth]{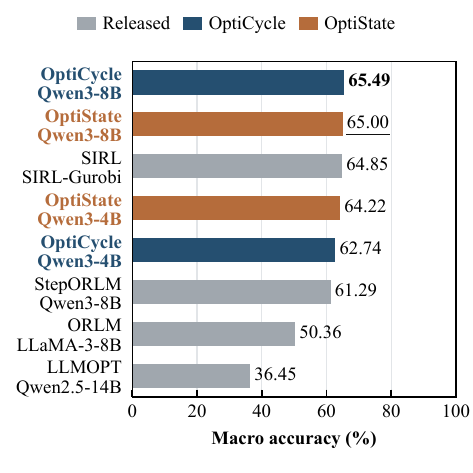}
\caption{Macro accuracy (\%) across nine benchmarks, ordered by accuracy. Blue and orange highlight OptiCycle and OptiState; gray denotes released models. Bold and underline mark the highest and second-highest values, respectively.}
\label{fig:released_models}
\end{figure}

Without any additional training or fine-tuning, our harnesses outperform
several post-trained \mbox{optimization} models in macro accuracy
(Figure~\ref{fig:released_models}). Both frozen Qwen3-8B configurations
exceed StepORLM, SIRL, ORLM, and LLMOPT.

\subsection{Selective Benefits of Formulation Structure}
\label{sec:results_structure}
Adding explicit formulation state produces \mbox{selective} gains
(Figure~\ref{fig:harness_gains}, left). The largest \mbox{improvements} occur on
\texttt{Qwen3.5-4B} and \texttt{Qwen3-1.7B}, while several other backbones show little
benefit or a decline relative to \textsc{OptiCycle}. This pattern is
not monotonic in model size: structured state helps some smaller
models substantially but does not reliably improve the smallest
variants. Its contribution is therefore more dependent on the
backbone than the gains from executable interaction.

\section{Analysis}
\label{sec:analysis}

\subsection{Where Interaction Helps Most}

Our results suggest that a simple execution-and-revision harness is
most useful when tasks \mbox{challenge} direct generation but remain
within the model's capabilities. ComplexOR illustrates this pattern:
\textsc{OptiCycle} substantially improves over Direct Generation on
both shared backbones. Gains are smaller on MAMO-easy for \texttt{Qwen3-4B}
and \texttt{Qwen3-8B}, where direct generation accuracy is \mbox{already} high. Interaction
also offers little \mbox{benefit} for some difficult model--task pairs,
such as \texttt{Qwen3-8B} on MAMO-complex
(Tables~\ref{tab:qwen3_benchmarks} and~\ref{tab:qwen35_benchmarks}).

These findings are consistent with a harness \mbox{helping} models make
better use of their existing capabilities, with limited headroom
on easier tasks and persistent capability constraints on harder
ones. Difficulty alone, however, does not determine the gains:
OptMath improves substantially despite low final accuracy,
showing that interaction can help even when reliable task
completion remains out of reach.

\subsection{Formulation State Helps Some Models and Tasks}

Explicit formulation state provides additional gains beyond
executable interaction, but its value \mbox{depends }on the backbone and
task. For \texttt{Qwen3-1.7B} and \texttt{Qwen3.5-4B}, \mbox{\textsc{OptiState}} \mbox{improves}
over \textsc{OptiCycle} on all nine \mbox{benchmarks}, indicating broad
benefits for these two models
(Tables~\ref{tab:qwen3_benchmarks} and~\ref{tab:qwen35_benchmarks}).

For \texttt{Qwen3-8B} and \texttt{Qwen3-14B}, the effects are mixed. Both models
improve on NL4Opt and ICML Competition but lose accuracy on
OptMath and ComplexOR. The small changes in macro \mbox{accuracy}
therefore mask opposing effects across individual benchmarks.
These results suggest that the value of explicit formulation
state should be assessed for each model and task, rather than
\mbox{assumed} to improve the execution loop uniformly.

\subsection{Performance Beyond the Qwen Family}
\label{sec:non_qwen_analysis}

\begin{figure}[t]
\centering
\includegraphics[width=\columnwidth]{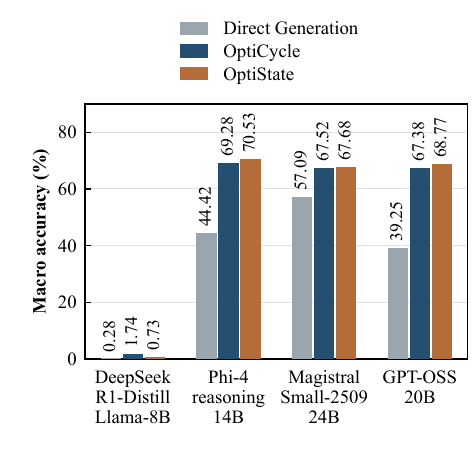}
\caption{Macro accuracy (\%) across nine benchmarks on non-Qwen models, including GPT-OSS-20B. Values are unweighted means of benchmark avg@$k$ scores.}
\label{fig:non_qwen_macro}
\end{figure}

Executable interaction also benefits non-Qwen models
(Figure~\ref{fig:non_qwen_macro}). \textsc{OptiCycle} raises macro
\mbox{accuracy} from 44.42\% to 69.28\% on \mbox{Phi-4-reasoning} and from
57.09\% to 67.52\% on Magistral-Small-2509; \textsc{OptiState}
adds 1.25 and 0.16 percentage points, respectively. On GPT-OSS-20B,
accuracy rises from 39.25\% (Direct Generation) to 67.38\%
(\textsc{OptiCycle}) and 68.77\% (\textsc{OptiState}), gains of
28.13 and a further 1.39 percentage points. \textsc{OptiCycle}
accounts for most of the gains on these three models.

DeepSeek-R1-Distill-Llama-8B remains below 2\%: 0.28\% with
Direct Generation, 1.74\% with \textsc{OptiCycle}, and 0.73\%
with \textsc{OptiState}. The benefits therefore vary across backbones.

\section{Conclusion and Future Work}
\label{sec:conclusion}

We studied optimization modeling as an \mbox{executable} inference task for frozen
language models. Across nine benchmarks and 11 Qwen \mbox{variants}, a lightweight
execution-and-repair harness \mbox{substantially} improves performance over
direct generation for models above 1B parameters and outperforms the
evaluated adapted architectures with \texttt{Qwen3-4B} and \texttt{Qwen3.5-9B}. Explicit formulation state
provides additional, model-dependent gains. The same workflow also
improves accuracy on Phi, Magistral, and GPT-OSS.

\paragraph{Future work.}
An important direction is a \mbox{\emph{self-evolving harness}} that improves its
\mbox{inference} workflow across tasks while keeping the \mbox{underlying} model frozen.
Rather than revising only a task's code or formulation, such a framework
could use historical execution trajectories to \mbox{propose} changes to prompts,
feedback \mbox{processing}, state \mbox{representations}, and repair-budget \mbox{allocation}.
\mbox{Candidate} harnesses could be evaluated on a separate \mbox{development} set and
retained when they\mbox{ improve }accuracy under explicit cost constraints.

\section*{Limitations}

While our harnesses improve optimization-modeling performance across the
evaluated benchmarks, the experiments compare complete workflows with
method-specific inference budgets. In particular, \textsc{OptiState}
adds both an initial formulation call and a revisable representation to
\textsc{OptiCycle}. We report recorded token consumption for the
Qwen3.5-9B comparison; matched-budget ablations, trajectory-level repair
analysis, and latency measurements can further clarify the contributions
and costs of individual components.

We have primarily evaluated 11 Qwen variants, with additional experiments
on DeepSeek-R1-Distill-Llama-8B, Phi-4-reasoning, Magistral-Small-2509,
and GPT-OSS-20B. Configuration revisions differ across model generations,
so these results characterize the evaluated model--harness settings.
The reported accuracies are point estimates without uncertainty
intervals; broader model coverage and uncertainty quantification would
further strengthen the empirical findings.

Finally, our evaluation uses objective-value or status agreement as a
common outcome-level measure across benchmarks. Extending verification
to decision-vector feasibility, constraint satisfaction, and formulation
equivalence would provide a more comprehensive assessment of modeling
correctness.

\section*{Ethical Considerations}

This work studies the use of language models for automated
optimization modeling. Incorrect formulations or solver
implementations may produce misleading solutions, especially
in consequential decision-making settings. Our evaluation
focuses on benchmark accuracy and does not establish
reliability for real-world deployment. Human verification
remains important before applying generated optimization
models. Additionally, repeated model inference and solver
execution incur computational costs.

\bibliography{related_work}

\clearpage
\appendix
\raggedbottom
\makeatletter
\setlength{\@fptop}{0pt}
\setlength{\@dblfptop}{0pt}
\makeatother
\section{Experimental Details}
\label{sec:experimental_details}

\paragraph{ICML Competition.}
This benchmark uses test problems from the \emph{ICML 2024 Challenges
on Automated Math Reasoning---Track 3: Automated Optimization
Problem-Solving with Code}~\cite{icml2024optimizationcompetition}.
We refer to it as ICML Competition throughout the paper and abbreviate
it as ICML Comp. in result tables.

\subsection{Models and Sampling}
\label{sec:sampling_details}
The model-scale comparison uses \texttt{Qwen3-0.6B, 1.7B, 4B, 8B}, and \texttt{14B};
\texttt{Qwen3.5-0.8B, 2B, 4B, 9B,} and \texttt{27B}; and \texttt{Qwen3.6-27B}. The nine-method
comparison uses \texttt{Qwen3-4B} and \texttt{Qwen3.5-9B}. Model weights remain fixed in
all evaluations. Configuration revisions differ across model generations.

Table~\ref{tab:sampling_details} lists the number of planned trajectories
per problem. Each model--method evaluation plans 17,852
trajectories over 2,634 problems. We use temperature 0.6, top-$p$ 0.95,
top-$k$ 20, and seed $42+j$ for sample index $j$. All model calls,
including individual stages of the baseline workflows, use a maximum
output length of 16,384 tokens. This per-call limit is shared across
experiments; total inference budgets still differ with the workflow's
number of calls.

\begin{table}[ht]
\centering

\small
\caption{Per-problem sampling counts used to estimate average accuracy.}
\label{tab:sampling_details}
\begin{tabular}{lr}
\toprule
Benchmark & Trajectories per problem ($k$) \\
\midrule
ComplexOR & 64 \\
IndustryOR & 16 \\
OptiBench & 4 \\
MAMO-easy & 4 \\
MAMO-complex & 8 \\
NLP4LP & 8 \\
NL4Opt & 8 \\
ICML Competition & 8 \\
OptMath & 8 \\
\bottomrule
\end{tabular}

\end{table}

\paragraph{Token accounting.}
For the Qwen3.5-9B comparison, we use the current evaluated runs and
exclude archived retries. Each method has 17,852 scored trajectories
with recorded input and output token counts. We sum input and output
tokens over the model calls within each trajectory, compute the mean
within each benchmark, and then average the nine benchmark means
without weighting. Recorded zero-token pairs are retained. This
metric measures recorded token consumption rather than monetary cost
or latency.

\paragraph{Qwen3.5 scoring coverage.}
Across the three methods, \texttt{Qwen3.5-0.8B}, \texttt{Qwen3.5-4B}, and the
\texttt{Qwen3.5-9B} subset each have all 53,556 scores. \texttt{Qwen3.5-2B} has 53,555
scores: one OptiState trajectory on OptMath has no recorded score.
\texttt{Qwen3.5-27B} has 53,553 scores: three OptiCycle trajectories remain
without recorded scores, affecting IndustryOR, MAMO-complex, and OptMath. The
\texttt{Qwen3.5-27B} results also include one scored trajectory with a timeout exception.
We assign zero to all missing scores and retain every planned
trajectory in the accuracy denominator. The $\dagger$ marks identify
affected evaluations, and all tables and figures use this scoring rule.

\subsection{Baseline Adaptations and Budgets}
\label{sec:baseline_details}
We adapt the baseline workflows from their papers and original code to
the shared OptAgentBench runner. Table~\ref{tab:workflow_budgets} summarizes the
implemented procedures and their budgets. The methods share data,
solver access, sampling, execution infrastructure, and grading within
each backbone comparison, but differ in the number and organization of
model calls. Table~\ref{tab:workflow_budgets} gives the call or step limits for each workflow.

\begin{table*}[!t]
\centering

\small
\setlength{\tabcolsep}{7pt}
\renewcommand{\arraystretch}{1.15}
\caption{Implemented workflows and method-specific inference budgets.}
\label{tab:workflow_budgets}
\begin{tabular}{@{}>{\raggedright\arraybackslash}p{0.18\linewidth}
>{\raggedright\arraybackslash}p{\dimexpr0.56\linewidth-2\tabcolsep\relax}
>{\raggedright\arraybackslash}p{\dimexpr0.26\linewidth-2\tabcolsep\relax}@{}}
\toprule
\textbf{Method} & \textbf{Adapted procedure} & \textbf{Budget} \\
\midrule
Direct Generation & Generate a complete solver program and execute it once. & One model call \\
\addlinespace[4pt]
CAFA & Use compact OR-specific prompting and examples. & One model call \\
\addlinespace[4pt]
Reflexion & Generate tests and an initial program, execute it, reflect on the observations, and revise. & At most one revision \\
\addlinespace[4pt]
Chain-of-Experts & A conductor selects three of seven expert types; a reducer combines their outputs. & Three experts and a reducer \\
\addlinespace[4pt]
OptiMUS & A manager schedules Formulator, Programmer, and Evaluator roles over structured state. & At most ten scheduling steps \\
\addlinespace[4pt]
AutoFormulation & Use UCT search over parameters, variables, the objective, equalities, and inequalities. & Eight search steps; branching factor three \\
\addlinespace[4pt]
ORMind & Run a Semantic Encoder, Formalizer, Compiler, Supervisor, and System-2 checker. & One runtime repair \\
\addlinespace[4pt]
\textsc{OptiCycle} & Generate and repair code in a persistent Python interpreter. & At most ten code turns \\
\addlinespace[4pt]
\textsc{OptiState} & Add an initial formulation and revisable five-element state to \textsc{OptiCycle}. & Formulation call plus at most ten code turns \\
\bottomrule
\end{tabular}

\end{table*}

\subsection{Execution and Grading}
\label{sec:grading_details}
Generated programs execute in Python with SciPy, PuLP, and PySCIPOpt.
The runner uses host subprocesses within temporary task directories;
hidden grader files are not copied into these directories. Baseline
inference used eight independent vLLM endpoints. Structured
JSON, routing, and compilation stages may use direct decoding, while
stages involving semantic analysis, search, or traceback debugging retain
reasoning, according to each harness's configuration.

For each benchmark, avg@$k$ is the percentage of correct trajectories
among all planned samples over all problems. Trajectories with missing
scores count as incorrect and remain in the denominator.
It is an estimate of average
single-trajectory accuracy, not the probability that at least one of
$k$ samples succeeds. Macro accuracy is the unweighted mean of these
benchmark accuracies. The grader compares numerical objective values
using absolute tolerance $10^{-2}$ and relative tolerance $10^{-4}$,
or checks agreement on \texttt{infeasible}/\texttt{unbounded} status.
Missing or unparseable answers count as failures. Grading is performed
after inference and is not used for search, reflection, or candidate
selection. 

\section{Benchmark-Level Results}
\label{sec:updated_details}
Tables~\ref{tab:qwen3_benchmarks} and~\ref{tab:qwen35_benchmarks}
report benchmark-level accuracy for all 11 model variants, including
\texttt{Qwen3-4B} and \texttt{Qwen3.5-9B}. Each model forms a separate block with methods
as rows and benchmarks as columns. Accuracy is measured by avg@$k$ (\%),
with sampling counts shown in the second header row; Macro is the
unweighted mean across the nine benchmarks. Bold and underlined values mark
the best and second-best results, respectively, among Direct Generation,
\textsc{OptiCycle}, and \textsc{OptiState} within
each model and column, including ties.
Sampling and grading follow
Appendix~\ref{sec:experimental_details}.

\begin{table*}[!t]
\centering

\footnotesize
\setlength{\tabcolsep}{2pt}
\renewcommand{\arraystretch}{1.12}
\caption{Qwen3 benchmark accuracy (\%). Model blocks compare the three inference interfaces.}
\label{tab:qwen3_benchmarks}
\begin{tabular*}{\textwidth}{@{\extracolsep{\fill}}l|c|c|c|c|c|c|c|c|c|c@{}}
\toprule
\raisebox{0.5\normalbaselineskip}{Method} & \textbf{\shortstack{Complex\\OR}} & \textbf{\shortstack{Industry\\OR}} & \raisebox{0.5\normalbaselineskip}{\textbf{OptiBench}} & \textbf{\shortstack{MAMO\\easy}} & \textbf{\shortstack{MAMO\\complex}} & \raisebox{0.5\normalbaselineskip}{\textbf{NLP4LP}} & \raisebox{0.5\normalbaselineskip}{\textbf{NL4Opt}} & \textbf{\shortstack{ICML\\Comp.}} & \raisebox{0.5\normalbaselineskip}{\textbf{OptMath}} & \raisebox{0.5\normalbaselineskip}{\textbf{Macro}} \\
\cmidrule(lr){2-11}
& avg@64 & avg@16 & avg@4 & avg@4 & avg@8 & avg@8 & avg@8 & avg@8 & avg@8 & Avg. \\
\midrule
\multicolumn{11}{c}{\textbf{\texttt{Qwen3-0.6B}}} \\
\midrule
Direct Generation & \textbf{0.00} & \textbf{0.12} & \textbf{0.04} & \underline{0.00} & \textbf{0.18} & \textbf{0.05} & \underline{0.00} & \textbf{0.00} & \textbf{0.15} & \textbf{0.06} \\
OptiCycle & \textbf{0.00} & \underline{0.06} & \underline{0.00} & \underline{0.00} & \underline{0.00} & \underline{0.00} & \underline{0.00} & \textbf{0.00} & \underline{0.00} & \underline{0.01} \\
OptiState & \textbf{0.00} & 0.00 & \textbf{0.04} & \textbf{0.04} & \underline{0.00} & \underline{0.00} & \textbf{0.05} & \textbf{0.00} & \underline{0.00} & \underline{0.01} \\
\midrule
\multicolumn{11}{c}{\textbf{\texttt{Qwen3-1.7B}}} \\
\midrule
Direct Generation & 2.17 & 2.69 & 3.55 & 0.96 & \underline{1.66} & 5.68 & 4.73 & 3.78 & \underline{1.13} & 2.93 \\
OptiCycle & \underline{4.43} & \underline{4.88} & \underline{15.50} & \underline{25.04} & 1.18 & \underline{21.75} & \underline{23.37} & \underline{22.90} & 0.53 & \underline{13.28} \\
OptiState & \textbf{17.10} & \textbf{10.94} & \textbf{30.21} & \textbf{48.77} & \textbf{2.61} & \textbf{42.61} & \textbf{45.11} & \textbf{42.44} & \textbf{1.28} & \textbf{26.79} \\
\midrule
\multicolumn{11}{c}{\textbf{\texttt{Qwen3-4B}}} \\
\midrule
Direct Generation & 29.08 & 33.88 & 54.38 & 80.06 & 27.61 & 72.88 & 72.45 & 72.62 & 9.41 & 50.26 \\
OptiCycle & \textbf{66.58} & \underline{40.06} & \underline{70.12} & \underline{85.70} & \underline{33.06} & \underline{74.64} & \underline{86.25} & \underline{85.18} & \textbf{23.04} & \underline{62.74} \\
OptiState & \underline{65.45} & \textbf{40.13} & \textbf{73.18} & \textbf{87.58} & \textbf{36.49} & \textbf{75.67} & \textbf{89.46} & \textbf{88.23} & \underline{21.76} & \textbf{64.22} \\
\midrule
\multicolumn{11}{c}{\textbf{\texttt{Qwen3-8B}}} \\
\midrule
Direct Generation & 46.61 & 40.56 & 59.38 & 85.93 & \textbf{37.62} & 73.81 & 81.03 & 80.34 & 16.79 & 58.01 \\
OptiCycle & \textbf{67.88} & \underline{42.56} & \underline{72.93} & \underline{87.77} & \underline{36.73} & \underline{75.52} & \underline{87.72} & \underline{86.19} & \textbf{32.08} & \textbf{65.49} \\
OptiState & \underline{64.50} & \textbf{42.69} & \textbf{73.22} & \textbf{88.34} & 36.67 & \textbf{76.86} & \textbf{89.46} & \textbf{88.78} & \underline{24.47} & \underline{65.00} \\
\midrule
\multicolumn{11}{c}{\textbf{\texttt{Qwen3-14B}}} \\
\midrule
Direct Generation & 12.59 & 14.56 & 15.87 & 27.22 & 15.28 & 19.21 & 22.39 & 22.47 & 8.13 & 17.53 \\
OptiCycle & \textbf{69.88} & \textbf{47.12} & \underline{73.88} & \textbf{88.31} & \textbf{38.80} & \textbf{78.00} & \underline{89.08} & \underline{88.23} & \textbf{29.59} & \textbf{66.99} \\
OptiState & \underline{65.89} & \underline{46.69} & \textbf{75.25} & \underline{87.38} & \underline{37.14} & \underline{75.88} & \textbf{92.34} & \textbf{90.85} & \underline{25.98} & \underline{66.38} \\
\bottomrule
\end{tabular*}

\end{table*}

\begin{table*}[!t]
\centering

\footnotesize
\setlength{\tabcolsep}{2pt}
\renewcommand{\arraystretch}{1.12}
\caption{Qwen3.5 and Qwen3.6 benchmark accuracy (\%), grouped by model. $\dagger$ marks evaluations with missing scores counted as failures.}
\label{tab:qwen35_benchmarks}
\begin{tabular*}{\textwidth}{@{\extracolsep{\fill}}l|c|c|c|c|c|c|c|c|c|c@{}}
\toprule
\raisebox{0.5\normalbaselineskip}{Method} & \textbf{\shortstack{Complex\\OR}} & \textbf{\shortstack{Industry\\OR}} & \raisebox{0.5\normalbaselineskip}{\textbf{OptiBench}} & \textbf{\shortstack{MAMO\\easy}} & \textbf{\shortstack{MAMO\\complex}} & \raisebox{0.5\normalbaselineskip}{\textbf{NLP4LP}} & \raisebox{0.5\normalbaselineskip}{\textbf{NL4Opt}} & \textbf{\shortstack{ICML\\Comp.}} & \raisebox{0.5\normalbaselineskip}{\textbf{OptMath}} & \raisebox{0.5\normalbaselineskip}{\textbf{Macro}} \\
\cmidrule(lr){2-11}
& avg@64 & avg@16 & avg@4 & avg@4 & avg@8 & avg@8 & avg@8 & avg@8 & avg@8 & Avg. \\
\midrule
\multicolumn{11}{c}{\textbf{\texttt{Qwen3.5-0.8B}}} \\
\midrule
Direct Generation & 0.00 & \underline{0.00} & 0.00 & 0.00 & \underline{0.00} & 0.00 & 0.00 & 0.00 & \underline{0.00} & 0.00 \\
OptiCycle & \textbf{0.43} & \textbf{0.31} & \textbf{2.73} & \textbf{8.82} & \underline{0.00} & \textbf{3.31} & \textbf{3.21} & \textbf{3.29} & \textbf{0.23} & \textbf{2.48} \\
OptiState & \underline{0.17} & \underline{0.00} & \underline{0.74} & \underline{2.80} & \textbf{0.06} & \underline{0.72} & \underline{0.82} & \underline{1.07} & \underline{0.00} & \underline{0.71} \\
\midrule
\multicolumn{11}{c}{\textbf{\texttt{Qwen3.5-2B}}} \\
\midrule
Direct Generation & 0.09 & 0.06 & 0.12 & 0.31 & 0.12 & 0.31 & 0.16 & 0.15 & 0.00 & 0.15 \\
OptiCycle & \textbf{5.03} & \textbf{4.31} & \textbf{11.78} & \textbf{27.07} & \textbf{1.01} & \textbf{16.32} & \textbf{17.45} & \textbf{16.31} & \textbf{0.90} & \textbf{11.13} \\
OptiState & \underline{4.25} & \underline{3.75} & \underline{11.20} & \underline{20.40} & \underline{0.89} & \underline{13.64} & \underline{16.68} & \underline{14.05} & \underline{0.83}$^{\dagger}$ & \underline{9.52}$^{\dagger}$ \\
\midrule
\multicolumn{11}{c}{\textbf{\texttt{Qwen3.5-4B}}} \\
\midrule
Direct Generation & 4.86 & 8.75 & 10.58 & 7.32 & 7.17 & 11.47 & 11.47 & 13.54 & 2.03 & 8.58 \\
OptiCycle & \underline{22.57} & \underline{18.25} & \underline{44.34} & \underline{60.39} & \underline{8.06} & \underline{55.22} & \underline{58.91} & \underline{55.58} & \underline{7.98} & \underline{36.81} \\
OptiState & \textbf{48.70} & \textbf{35.38} & \textbf{66.78} & \textbf{80.90} & \textbf{20.91} & \textbf{71.49} & \textbf{86.20} & \textbf{82.53} & \textbf{15.06} & \textbf{56.44} \\
\midrule
\multicolumn{11}{c}{\textbf{\texttt{Qwen3.5-9B}}} \\
\midrule
Direct Generation & 34.64 & 27.50 & 51.20 & 69.13 & 22.22 & 62.86 & 71.79 & 67.53 & 11.37 & 46.47 \\
OptiCycle & \underline{64.06} & \underline{43.62} & \underline{73.68} & \textbf{89.00} & \underline{31.93} & \textbf{77.79} & \underline{87.66} & \underline{87.74} & \underline{31.40} & \underline{65.21} \\
OptiState & \textbf{65.54} & \textbf{47.19} & \textbf{75.33} & \underline{86.66} & \textbf{38.39} & \underline{77.58} & \textbf{92.66} & \textbf{90.49} & \textbf{33.81} & \textbf{67.52} \\
\midrule
\multicolumn{11}{c}{\textbf{\texttt{Qwen3.5-27B}}} \\
\midrule
Direct Generation & 53.12 & \underline{50.44} & 68.22 & 85.77 & 44.79 & 76.14 & 88.04 & 84.66 & 34.86 & 65.12 \\
OptiCycle & \textbf{68.84} & 49.38$^{\dagger}$ & \underline{77.19} & \textbf{88.31} & \underline{48.05}$^{\dagger}$ & \underline{77.79} & \underline{89.73} & \underline{89.48} & \underline{52.11}$^{\dagger}$ & \underline{71.21}$^{\dagger}$ \\
OptiState & \underline{68.49} & \textbf{56.50} & \textbf{78.51} & \underline{87.58} & \textbf{52.90} & \textbf{78.05} & \textbf{93.48} & \textbf{90.95} & \textbf{52.86} & \textbf{73.26} \\
\midrule
\multicolumn{11}{c}{\textbf{\texttt{Qwen3.6-27B}}} \\
\midrule
Direct Generation & 58.94 & 49.06 & 71.61 & 83.78 & \underline{46.56} & \underline{76.29} & 89.46 & 85.88 & 36.67 & 66.47 \\
OptiCycle & \textbf{73.70} & \textbf{57.06} & \underline{76.61} & \underline{88.65} & \textbf{48.46} & 76.08 & \underline{90.38} & \underline{90.40} & \textbf{50.98} & \textbf{72.48} \\
OptiState & \underline{73.44} & \underline{57.00} & \textbf{78.88} & \textbf{88.73} & 46.09 & \textbf{76.76} & \textbf{90.82} & \textbf{90.61} & \underline{48.72} & \underline{72.34} \\
\bottomrule
\end{tabular*}

\end{table*}

\end{document}